# Beauty is in the AI of the beholder: MLLMs systematically overrate facial attractiveness

Santiago Grandas, Juan Sebastian Cely-Acosta, Mohit Mendiratta, Shafee Hassan, Macken Murphy

Qoves Inc., Wilmington, DE, United States

CORRESPONDING AUTHOR Santiago Grandas santiago@qoves.com

## Ethics statement:

Qoves Inc. (Wilmington, DE, United States) holds institutional and ethical responsibility for this research. This study did not require institutional review board approval because no new data were collected from human participants. We employ only secondary analysis of the Face Research Lab London Set (DeBruine & Jones, 2017), a publicly available face image dataset released under a CC BY 4.0 license, together with generated ratings from four publicly available AI models. All individuals in the dataset gave signed consent for their images to be 'used in lab-based and web-based studies in their original or altered forms and to illustrate research (e.g., in scientific journals, news media, or presentations)' (DeBruine & Jones, 2017).

## Note from the authors:

This is version one of a preprint. Findings presented here have not been peer reviewed and should be treated accordingly, both by the press and by the academy.

## Conflict of interest statement:

This study was run by QOVES, a company that sells personalized beauty advice. Santiago Grandas, Juan Sebastian Cely-Acosta, and Mohit Mendiratta are all employed full-time as scientists by Qoves Inc. or its affiliates; Grandas and Mendiratta hold equity in the company. Shafee Hassan is the founder and chief technical officer of Qoves Inc. and holds equity in the company. Macken Murphy works part-time as a paid contractor of Qoves Inc. and holds equity in the company. The authors declare no other competing interests.

# Beauty is in the AI of the beholder: MLLMs systematically overrate facial attractiveness

Santiago Grandas, Juan Sebastian Cely-Acosta, Mohit Mendiratta, Shafee Hassan, Macken Murphy

Qoves Inc., Wilmington, DE, United States

Beauty assessments from Multimodal Large Language Models (MLLMs) are increasingly popular amongst users, companies, and aestheticians. This raises the question of whether these AI models can accurately reflect human judgments of attractiveness. In a pre-registered exploratory study, we compared the attractiveness ratings of 2,513 human participants to four widely used commercial AI models: Claude, Gemini, GPT, and Grok. Results showed that MLLMs systematically rate faces more favorably and within a narrower range than humans and, at the time of study, do not reproduce human ratings in absolute terms. However, MLLMs exhibit strong correlations with human attractiveness judgments, accurately tracking the rank-ordering of faces. MLLMs may judge faces by different cues than humans; only face age was a predictor of facial attractiveness in both humans and MLLMs, with inconsistent patterns across models for ethnicity and gender. AI models strongly agree with one another, except for Grok, which also showed the lowest agreement with humans. Our findings suggest that while they may be able to approximate rank-orderings of human attractiveness, current off-the-shelf commercial MLLMs systematically overrate the beauty of human faces.

*"Beauty is no quality in things themselves: It exists merely in the mind which contemplates them; and each mind perceives a different beauty" — David Hume*

## 1. Introduction

The proliferation of multimodal large language models (MLLMs) and the age-old desire to quantify beauty have led everyday users, companies, and aesthetic practitioners to utilize AI to evaluate human faces (Goshtasbi et al., 2024; Haider et al., 2025; Sternlicht, 2024). MLLM attractiveness ratings are becoming increasingly popular: as of 2026, several dozen AI-powered apps and websites claim to evaluate users' looks and attractiveness, with some reaching over 7 million downloads (Sternlicht, 2024). Accessibility, low cost, and, crucially, an illusion of objectivity have encouraged users of general-purpose MLLMs to ask these systems to evaluate their own and others' facial attractiveness (Goshtasbi et al., 2024). But while inexpensive and quick, these assessments of attractiveness are not objective; rather, they simply rely on a new and understudied class of judges.

While MLLMs have demonstrated strong image-processing capabilities (Liu et al., 2024), leading models still struggle with basic face-understanding tasks. In general, they perform poorly on fine-grained facial perception tasks that require feature extraction and spatial understanding of facial structure, and show low accuracy on tasks such as age and race estimation (Narayan et al., 2026). Nevertheless, studies comparing MLLM and human judgments on more holistic, social facial tasks, such as facial emotion identification (Nelson et al., 2025) and social judgments of face images (Hausladen et al., 2024), report overall high consistency with human raters.

To date, very few studies have compared human and MLLM perceptions of facial attractiveness. Preliminary findings suggest a strong correlation between humans and AI systems, including frontier models (GPT; Kramer, 2025) and specialized AI-based facial analysis websites (Goshtasbi et al., 2024). However, current findings are subject to scrutiny due to methodological considerations, which we discuss in detail. So, while people increasingly rely on these systems as stand-ins for human judgments of facial attractiveness, we still do not understand whether or to what degree these systems judge facial attractiveness like humans.

In this study, we attempt to answer this question by comparing four off-the-shelf frontier MLLMs' ratings of attractiveness against an existing dataset of human ratings to understand (i) the extent to which MLLM attractiveness ratings correspond to those provided by human raters, (ii) the extent to which four MLLMs' attractiveness ratings correspond to each other, (iii) whether MLLMs and humans show different patterns of association between face characteristics (sex, age, ethnicity) and attractiveness ratings, and (iv) whether MLLMs agree more or less with specific human rater subgroups defined by sex, age, and sexual preference (e.g., do MLLMs perceive facial attractiveness more similarly to men or women?).

### 1.1 Human perceptions of facial attractiveness

In order to compare MLLM perceptions of facial attractiveness with human perceptions of facial attractiveness, we must first review how humans perceive facial attractiveness. Facial attractiveness is a psychological phenomenon; it does not exist in the physical world, but in the minds of observers. Until recently, facial attractiveness perceptions were exclusively studied in human subjects, and so, since the 1970s (Dion et al., 1972), it has been primarily psychologists (and, to a lesser extent, other human social scientists) who have studied the science of human beauty perceptions. This topic is now well-explored, but complex, with the extant literature simultaneously evidencing strong cross-cultural consensus alongside differences between populations, between individuals within populations, as well as within individuals across time and contexts (Jenkins et al., 2011; Langlois et al., 2000; Leder et al., 2016; Rhodes, 2006).

Humans broadly agree with each other in their facial attractiveness judgments (Fink et al., 2007; Germine et al., 2015; Langlois et al., 2000; Perrett et al., 1999; Rhodes, 2006). For instance, studies by Leder et al. (2016) and Hönekopp (2006) both used variance component analysis in random-effects models and estimated that shared taste accounts for roughly 60% of the variance in attractiveness ratings, with individual taste accounting for the remaining 40%. Simply put, there is substantial consensus between individuals on what they find attractive (shared taste), but there is clearly significant variation from person to person (individual taste). Research from Langlois et al. (2000) and Rhodes (2006) suggests that agreement is particularly high when considering the extremes: bluntly, it is unlikely for some people to rate a face as very unattractive while others perceive it as highly attractive. So, while beauty is to some extent in the eye of the beholder, a substantial portion (perhaps most) of what makes faces attractive or unattractive is explicable by the qualities of the faces themselves, rather than the qualities of the minds that perceive them.

What, exactly, are the facial qualities that predict attractiveness? So far, scientists appear to have reached a broad consensus that the following facial qualities contribute to a beautiful appearance: averageness (defined as mathematically average trait values; Langlois & Roggman, 1990; Valentine et al., 2004), symmetry (Jones & Jaeger, 2019; Little et al., 2007; Perrett et al., 1999), cues to health (e.g., moderate facial adiposity; Coetzee et al., 2012; de Jager et al., 2018; Jones, 2018), cues to youth (i.e., the apparent age of a face; Foos & Clark, 2011; Korthase & Trenholme, 1982; Porcheron et al., 2017) when paired with cues to sexual maturity, facial neoteny in women (juvenile-like traits such as large eyes; Foos & Clark, 2011; Cunningham, 1986; Furnham & Reeves, 2006), facial femininity in women (Lee et al., 2025; Pavlovič et al., 2021; Perrett et al., 1998), and "Goldilocks dimorphism" in men (i.e., not too feminine, not too masculine; Cunningham et al., 1990; Scott et al., 2014). Relevantly to this study, female faces are generally perceived as more attractive than male faces by both men and women, a robust finding recently formalized as the "gender attractiveness gap" (Wassiliwizky et al., 2026).

Generally, these predictors of attractiveness have been interpreted as evolved in origin (Buss, 1995; Rhodes, 2006; Rhodes et al., 2007; Scheib et al., 1999), as they are hypothesized to signal traits that would be valuable in mates throughout the animal kingdom. For instance, facial symmetry has been associated with successful gene expression (e.g., Fink & Penton-Voak, 2002; A. L. Jones & Jaeger, 2019), and youthfulness has been linked to fertility (e.g., D. Jones, 1995; Thornhill & Gangestad, 1999). While this is reasonable, we must be cautious, as what is considered attractive varies with culture to a degree that would surprise many Westerners. For example, in the USA, straight, square teeth are most attractive, and one might assume this is due to an innate, evolved psychological preference for functional dentition. However, among the Aka (where shaving the teeth into points is common), square teeth are not preferred. As one Aka woman named Nali put it: "If you are a man or a woman searching for a wife or husband and you see someone who does not have pointed teeth, you say, 'You there, you are like a chimpanzee. You have big teeth like a chimpanzee! I do not want you.'" (Fancher, 2013; Hewlett, 2013).

Some of this apparent cultural variation may, in time, prove to be an adaptive response to the local ecology. For instance, it's plausible that cross-cultural differences in preferences for facial adiposity (thin faces versus chubby faces) could be downstream of differences in local resource abundance (Cazzato et al., 2022). Similarly, while "personal taste" in beauty is often conceptualized as though it is random noise (e.g., Wolf, 1991) (and, to some extent, it may well be; Scott et al., 2014) as Little & Perrett (2002) argue, these differences in taste are often predictably influenced by the individual's personal history and context. That is, much of the variation in beauty preferences that appears to be arbitrary may, in fact, not be.

The above-mentioned predictors and demographic effects have been well-explored in human raters. However, the well-documented variation across individuals and populations (regardless of source) poses a challenge to the "objectivity" and, indeed, the functional accuracy of AI beauty assessment, which may be tuned to mimic the taste of a subset of humans within a specific context.

### 1.2 MLLM perception of faces

Multimodal large language models (MLLMs) are general-purpose AI assistants trained on a vast amount of text and image data. Although frontier models (e.g., ChatGPT, Claude, Gemini) are generally understood to be built on transformer-based architectures (Vaswani et al., 2017), the specific architectural details, training data, and post-training procedures of each are proprietary and not publicly disclosed. Research on open-source vision-language models, however, offers a reasonable approximation of how these systems process visual input and what shapes their outputs (Radford et al., 2021; Schuhmann et al., 2022).

A defining feature of modern MLLMs is that they learn from image-text pairs scraped from the internet, typically on the order of hundreds of millions to several billion examples (Radford et al., 2021; Schuhmann et al., 2022). During pretraining, the model is taught to align the representation of an image with the representation of its paired caption, an approach popularized by the CLIP framework (Radford et al., 2021) and now foundational to most commercial models. Simply put, MLLMs do not perceive images in isolation; rather, they interpret visual features through the language humans have already used to describe them. A face is not simply encoded as a configuration of pixels, but as something co-occurring with accompanying captions, alt-text, and surrounding prose that appeared alongside it online.

The training nature of AI has two implications for the present study. First, AI judgments of faces are filtered through a linguistic prior; therefore, it is expected that ratings of attractiveness will at least partially reflect the verbal descriptions that have historically accompanied similar-looking faces in the training corpus. For instance, Hausladen et al. (2024) demonstrated that CLIP makes human-like social judgments from faces through this mechanism. Second, the image-text pairings underlying these models are demographically and culturally skewed, reflecting a westernized web landscape that is dominated by English language and stock and celebrity imagery (Birhane et al., 2021; Hausladen et al., 2024). This means that whatever consensus or bias is encoded in those captions is inherited by the model. Where a human attractiveness rating reflects a population-level average across embodied observers, an MLLM rating reflects an average across the descriptions of its training data.

It is worth noting that correspondence between AI and human judgments should not be mistaken for evidence that MLLMs perceive faces the way embodied human observers do. Empirical support for AI-human alignment is genuinely mixed. Some studies find statistical AI-human correspondence: AI models trained on brain activity from one sensory modality, such as language, can predict brain activity from another modality, such as vision. This finding suggests that language and vision share aspects of underlying conceptual structure in both artificial models and the human brain (Tang et al., 2023). Further, GPT-4V's annotations of social features from images and video correlate with human annotations at both the behavioral and neural level (Santavirta et al., 2025). But other work complicates the picture: multimodal training does not straightforwardly bring model representations closer to human experiential or perceptual norms, and in some cases reduces alignment relative to language-only models (Bavaresco & Fernández, 2025), and methodological critiques caution that some previously reported correspondences between model representations and human neural data reflect non-robust methods and overlooked confounds as opposed to genuine shared computation (Hadidi et al., 2026). Taken together, this suggests that any correspondence between MLLM and human attractiveness ratings is best understood as alignment in output, arising from a shared reliance on the linguistic descriptions historically attached to similar faces, instead of as evidence of a shared perceptual process.

Recent studies suggest that MLLMs' ratings of attractiveness may reflect not only human judgments of facial attractiveness, but also human preferences about rating behavior itself, as models appear to manifest a human-like social desirability bias in its outputs. For instance, work by Salecha and colleagues (2024) revealed that MLLMs consistently skew their Big Five personality scores towards the socially desirable ends of the trait dimensions. Similarly, in a research quality evaluation task, Thelwall (2024) found that MLLMs never awarded the lowest score to any of the reviewed papers. These biases may be explained by the well-documented phenomenon of sycophancy (Sharma et al., 2025), defined as the tendency of MLLMs to give agreeable, non-confrontational responses that match the user's belief over truthful ones, as a result of Reinforcement Learning from Human Feedback (RLHF) during their fine-tuning.

Following the rapid development and accessibility of AI, researchers in the fields of computer science (Hausladen et al., 2024; Wang et al., 2024), psychology (Kramer, 2025), and cosmetic surgery (Goshtasbi et al., 2024) have begun to explore the ability of these models to process facial images and assess them on a wide array of traits. These initial explorations into machine beauty psychology suggest MLLMs may track human judgment to a certain extent, as we will review in this section. However, this nascent literature is sparse: only a handful of studies have directly compared human judgments of social traits with those of AI models, and even fewer have examined attractiveness in particular.

Kramer's (2025) set of experimental studies stands out as an exception by directly comparing ChatGPT to human raters on judgments of attractiveness, dominance, and trustworthiness as inferred from facial photographs. Their results suggest that AI responses align with human judgments and even exhibit an attractiveness halo effect, a finding further supported by the work of Gulati and colleagues (2025). Kramer's results (2025) are partly based on forced-choice paired comparisons between faces drawn from the extremes of each trait. Although this approach demonstrates MLLMs can correctly discriminate between the low- and high-end of a trait (much like humans do), it fails to capture absolute agreement and rating differences when faces lie in the middle of the distribution.

A second experiment (Kramer, 2025, Study 1b) expanded on this by comparing individual ratings on a smaller sample. However, the results showed that ChatGPT test-retest reliability was lower than that observed across human raters, and the author explicitly flagged uncertainty about the appropriate baseline for AI consistency. This suggests that researchers should consider the non-deterministic nature of MLLMs and use consistent, stable measurements, as we acknowledge in our Methods section.

Rather than testing frontier MLLMs, some researchers have instead examined AI-based facial attractiveness websites. A study by Goshtasbi et al. (2024) compared the attractiveness ratings of an expert focus group (head surgery residents) to ratings provided by five AI-powered facial rating websites. Their facial stimuli were synthetic (i.e., AI-generated) and exclusively white and female. Although all five websites gave consistently higher ratings to all faces compared to human raters, the authors report a strong average AI-to-human correlation in attractiveness ratings (Goshtasbi et al., 2024).

In a slightly different vein, Haider et al. (2025) examined whether MLLMs could evaluate qualitative facial characteristics (e.g., skin type, fat distribution) and quantitative ratios (e.g., neoclassical canons) on 15 synthetic faces. Their study evaluated four frontier models (ChatGPT-4o, ChatGPT-4, Gemini 1.5 Pro, Claude 3.5 Sonnet) against three expert surgeons and manual measurements. The authors reported inconsistent reliability of MLLM judgments across faces and features, although some models displayed strong agreement with human raters on specific qualitative features (Haider et al., 2025). In parallel with our study, Haider and colleagues (2025) compare frontier MLLMs to humans, although their focus is on objective facial properties rather than the subjective aesthetic evaluation assessed here. Further, we test MLLMs' judgments against a large sample of raters, as opposed to a small number of experts.

Taken together, while preliminary results suggest that MLLMs do reflect human judgment, the extent to which they agree on attractiveness ratings remains unclear given the scarcity of research and the above-mentioned methodological limitations. Further, more nuanced questions, such as whether and to what degree MLLM ratings more closely approximate different subgroups of human raters (e.g., men versus women), and how facial characteristics (e.g., age, gender) shape MLLM attractiveness judgments, remain entirely unexplored.

### 1.3 Current study

This exploratory study investigates the agreement in facial attractiveness judgments between four MLLMs (GPT, Gemini, Claude, and Grok) and human raters, testing both the similarities and the differences in their ratings. We selected these four models because we intended to capture AI behavior for the average user and, at the time of writing, they are the four most widely used general-purpose AI products, ranked by monthly active users (Sensor Tower, 2026). Critically, we test these models as a typical user would encounter them, using their default, zero-shot configuration. Our study therefore describes the observed behavior of off-the-shelf frontier MLLMs in everyday use. We use a publicly available (yet unexplored for this purpose) dataset of 102 face photographs that had already been rated by a sample of 2513 human participants. Using a novel methodological approach, we obtained independent and stateless ratings for each face, via API calls, from all four MLLMs.

The current study seeks to explore both the agreement between MLLMs and humans (RQ1) and across different MLLMs (RQ2). Finally, we evaluate the predictive power of face-level variables (e.g., age) in ratings of attractiveness across MLLMs and humans (RQ3), and explore whether MLLMs show differential agreement with specific subgroups of raters (RQ4).

The present study has four main contributions. First, we include a stringent measure of absolute agreement by computing ICC, in addition to correlational measures, allowing us to uncover systematic biases in ratings across sources beyond rank-based tests. Second, we include four distinct and widely-used frontier models, thereby measuring agreement among models and between humans and specific MLLMs. Third, it employs non-synthetic human facial photographs rated by a large sample of 2513 raters. Finally, we provide preliminary evidence on face-level predictors of attractiveness in MLLMs, and on differential agreement between AI models and human rater subgroups.

## 2. Methods

The research questions, inclusion and exclusion criteria, and analysis plan for the current study were preregistered on the Open Science Framework (OSF; https://osf.io/7283f/overview?view_only =8da182 3cb322422e921e0fa86a8adb49) prior to data collection. Analysis scripts can be retrieved from: https://osf.io/kdvau/files/exkph. Deviations from the original plan are reported in the sections where they apply.

### 2.1 Stimuli and Human Ratings

Human attractiveness ratings (*n* = 2513) were obtained from the Face Research Lab London Set (DeBruine & Jones, 2017), a publicly available face image dataset released under a CC BY 4.0 license. We included the full set of 102 neutral, front-facing facial images: 48% female and 52% male, with a racial/ethnic composition of 67.6% White, 12.7% Black, 9.8% West-Asian, 8.8% East-Asian, and 1.0% mixed East-Asian/White. All 102 face images were 1350 × 1350 px JPEGs, normalized by the original dataset authors for pose, lighting, background, and framing (DeBruine & Jones, 2017). The mean age of the depicted individuals was 27.72 years (*SD* = 7.11). Similarly, we included the attractiveness ratings of all 2513 human raters[1]: 61.8% female, 38.0% male, and 0.1% intersex. Raters reported their sexual orientation as attracted primarily to men (56.2%), women (33.8%), either (9.6%), or neither (0.3%). Mean age of the raters was 26.72 years (*SD* = 10.04).

### 2.2 MLLM Ratings

Attractiveness ratings were collected from four widely used commercial MLLMs: GPT-5.3, Claude Sonnet 4, Gemini 2.5 Flash, and Grok 4.20[2]. We use the base, off-the-shelf generalist models without specific fine-tuning, reflecting their readily available, widely used capabilities. Ratings were obtained via independent, stateless Application Programming Interface (API) calls to minimize any influence of conversational memory or context. Each API call presented a single face image alongside the rating prompt, and the AI model returned a single numerical rating. In this framework, each face is rated in a zero-shot manner, in complete isolation, with no exposure to other faces in the set, previous context, or memory. All API calls were made with temperature set to 1.0 and with tool use, web search, and reasoning/thinking modes disabled. These settings approximate the default configuration encountered by a typical user of each model's standard chatbot interface, while preventing models from augmenting their responses with external information, thus capturing each model's internal representations of attractiveness rather than information retrieved at inference time.

[1]Although we pre-registered no exclusions, twelve (12) human raters whose responses were identical across all 102 faces (zero variance) were excluded from some analyses.

**2.2.1 Prompting.** In order to extract usable and standardized outputs from the AI models, prompt engineering was an important component of the study design. We take the rating scale and endpoints directly from the Face Research Lab London Set (1 = "much less attractive than average"; 7 = "much more attractive than average"). The exact item shown to human raters is not reported, however. We therefore adapted the question wording from DeBruine and colleagues (2007, Experiment 5), a study conducted by the same lab using the same 1-7 scale. The final user prompt given to the AI models was:

*"Please rate how attractive you think this face is by typing a number 1 (much less attractive than average) to 7 (much more attractive than average)".*

Pilot tests revealed that some models refused to provide attractiveness ratings when prompted as described above. A minimal system prompt was therefore included for task compliance. This is a limitation shared across the AI face-perception literature, with both Kramer (2025) and Haider et al. (2025) reporting reframed prompts to overcome refusals. The exact wording we used for our system prompt was as follows:

*"Your task is to provide numerical ratings of face images on standard psychological scales. Respond only with the number."*

Pilots revealed the added system prompt was effective in getting models to provide numerical ratings. Finally, we addressed technical API failures by retrying the API call up to 10 times with increasing time delays (as preregistered). Explicit refusals, as opposed to technical errors, were not re-attempted, coded as NA, and reported explicitly. Refusals were rare (10 of 6324 API calls) and occurred exclusively with Claude, which refused 10 of its 612 calls, concentrated on three faces[3]. Finally, all pilot data were discarded and not used in the main analyses.

**2.2.2 Convergence.** Large Language Models (LLMs) are stochastic models that do not provide identical responses to identical inputs every time. The implication for our study is that the same prompt may elicit slightly different ratings from the same AI model. With the aim of reflecting a "true" rating from each AI model, we obtained a highly stable output for each face by averaging that face's rating across multiple runs, with each run comprising one set of ratings for all 102 faces. To determine the required number of runs we employed the following pre-registered procedure: 1) Conducted a pilot of 20-30 runs per model; 2) computed the Intraclass Correlation Coefficient, ICC(1,1) treating each run as an independent rater; and 3) applied the Spearman-Brown formula to find the required number of runs per model needed to achieve a reliability of at least .95. This methodological contribution allows us to obtain a highly reliable estimate of the model's stable output for each face. In the pilot, single-run reliability was ICC(1,1) = .79 (Claude), .79 (GPT), .70 (Gemini), and .32 (Grok), requiring 6, 6, 9, and 41 runs, respectively, for main data collection.

**2.3 Data Analysis**

To address agreement between AI and human ratings (RQ1), and across models (RQ2), we drew from Kramer and colleagues (2025) and first computed pairwise correlation coefficients (Pearson and Spearman) between every possible pair of raters and then averaged these values, obtaining a metric that reflects the agreement between the attractiveness ratings of any given pair in our sample. Additionally, we computed ICC(2,1) as a measure of absolute agreement.

Although the calculation of Spearman's ρ was not specified in the pre-registration, we included it because the rating data violated the distributional assumptions underlying Pearson's r: the human ratings were discrete integer responses on a 1-7 scale, precluding interval-level measurement, and preliminary normality checks indicated departures from normality for both the human and MLLM rating distributions (see Results). We therefore report Spearman's ρ and the Intraclass Correlation Coefficient (ICC(2,1)) as the primary measures for RQ1 and RQ2 and provide Pearson's r results in the supplementary material.

Second, to measure how much a particular subject agreed with the rest of the group, we used the "leave one out" approach, whereby the ratings of all individual participants were correlated with the mean of the remaining participants.

To investigate the effect of facial characteristics across humans and MLLMs (RQ3), we fitted a single multilevel model using the lme4 package (Bates et al., 2015). The model regressed per-source face-mean attractiveness ratings on rater source (humans, Claude, GPT, Gemini, Grok), face characteristics (face gender, face age, and face ethnicity), and their interactions, with a random intercept for face. Finally, we computed estimated marginal means and within-source contrasts using the emmeans package. Although we contemplated separate linear regressions in our preregistration, we concluded that a single multilevel model with post hoc tests provided the same information.

Lastly, to examine whether MLLMs showed differential agreement across human rater subgroups (RQ4) we took each rater's individual agreement with each MLLM from RQ1 (their Pearson's r, Spearman's ρ, and ICC(2,1) across the faces) and averaged these coefficients within rater subgroups, retaining subgroups with at least 20 raters. We first did this for each rater demographic on its own (rater sex, rater age, and rater sexual preference), and then for more granular rater profiles formed by crossing these variables (e.g., males aged 17–22[4] with a preference for females), ordering subgroups from highest to lowest agreement to observe any differences in rating agreement across human subgroups and MLLMs. As an additional inferential test, we fitted five multilevel models (one for each AI model and one for the pooled-AI rating). Each model regressed the individual human rating on the corresponding AI's mean rating and included the main effects and interactions with all rater-level variables (rater sex, rater age, and rater sexual preference), with random intercepts for rater and face and a random slope for the AI rating across raters; sparse rater categories (intersex and no stated preference) were excluded.

These exclusions, the 20-rater minimum for subgroups, and the categorization of rater age into three groups were analytic decisions

[2]The exact models used were as follows: GPT-5.3 (model ID: gpt-5.3-chat-latest), Claude Sonnet 4 (model ID: claude-sonnet-4-20250514), Gemini 2.5 Flash (model ID: gemini-2.5-flash), Grok 4.20 (model ID: grok-4.20-0309-non-reasoning)

[3] Claude refused to rate faces 062_03, 038_03, and 135_03. These responses were coded as NA, and Claude's mean ratings for the affected faces were computed over the remaining valid runs.

[4] Being the only continuous variable, age was categorized into three groups to keep the cells balanced.

**Figure 1.** *Frequency distributions by source*

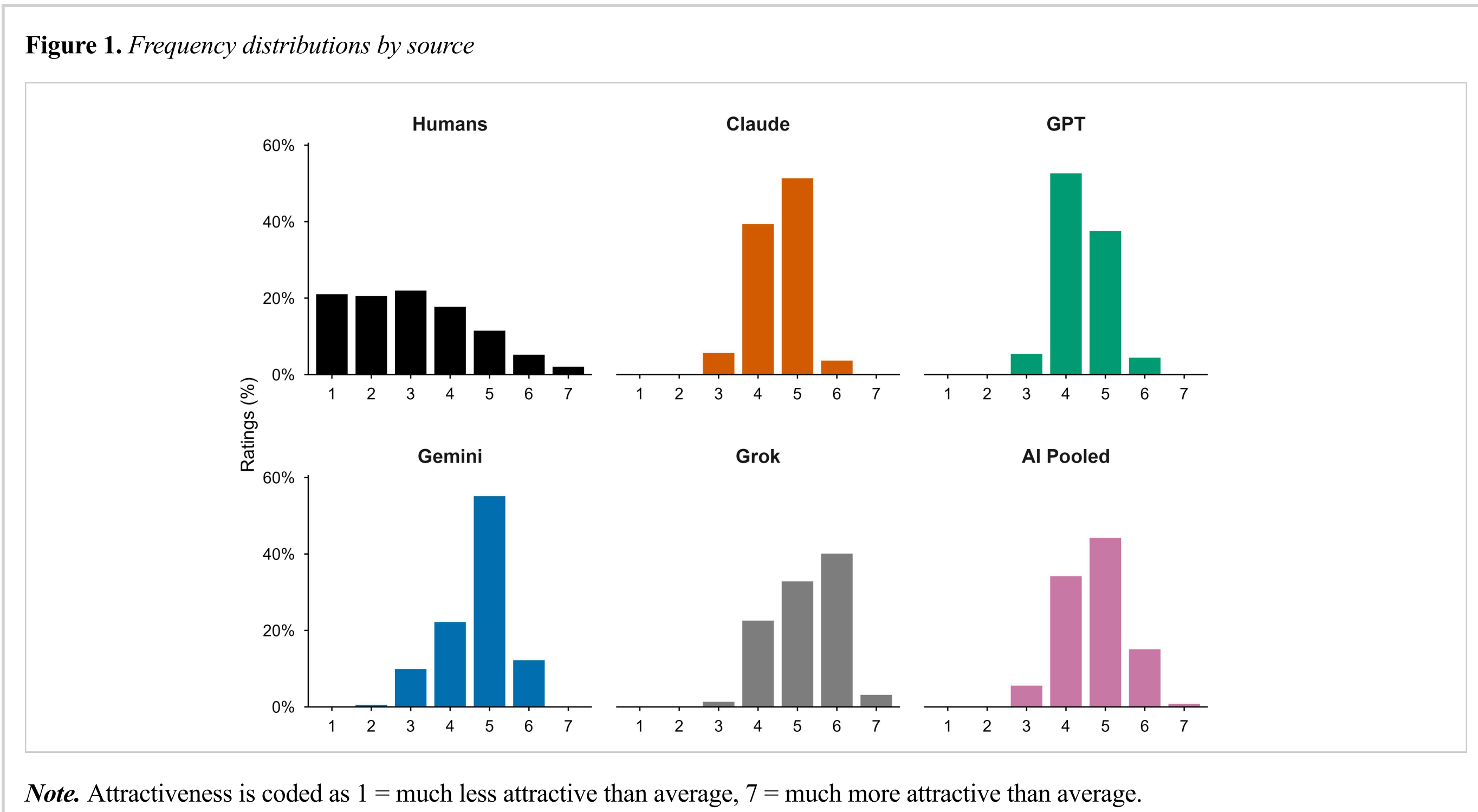

***Note.*** Attractiveness is coded as 1 = much less attractive than average, 7 = much more attractive than average.

not specified in the preregistration.

This study did not require institutional review board approval, as only secondary data and AI outputs were employed. Specifically, the study employs secondary analysis of the Face Research Lab London Set (DeBruine & Jones, 2017), a publicly available face image dataset released under a CC BY 4.0 license.

## 3. Results

We averaged all rater sources' ratings to obtain an overall mean attractiveness rating score for each of the MLLMs and for human raters. Additionally, we averaged all MLLM responses to produce a unified "pooled-AI" group. The average attractiveness ratings per source were as follows: Grok gave the highest mean rating ($M$ = 5.21, $SD$ = 0.87), followed by Gemini ($M$ = 4.69, $SD$ = 0.83), Claude ($M$ = 4.53, $SD$ = 0.66), and GPT ($M$ = 4.41, $SD$ = 0.66). Humans gave the lowest average rating ($M$ = 3.02, $SD$ = 1.57), which was descriptively lower than the average AI-model rating (AI-pooled, $M$ = 4.71, $SD$ = 0.55). The frequency distributions of attractiveness ratings across rater sources are reported in Figure 1.

**Figure 2.** *Pairwise agreement between MLLMs and individual human raters*

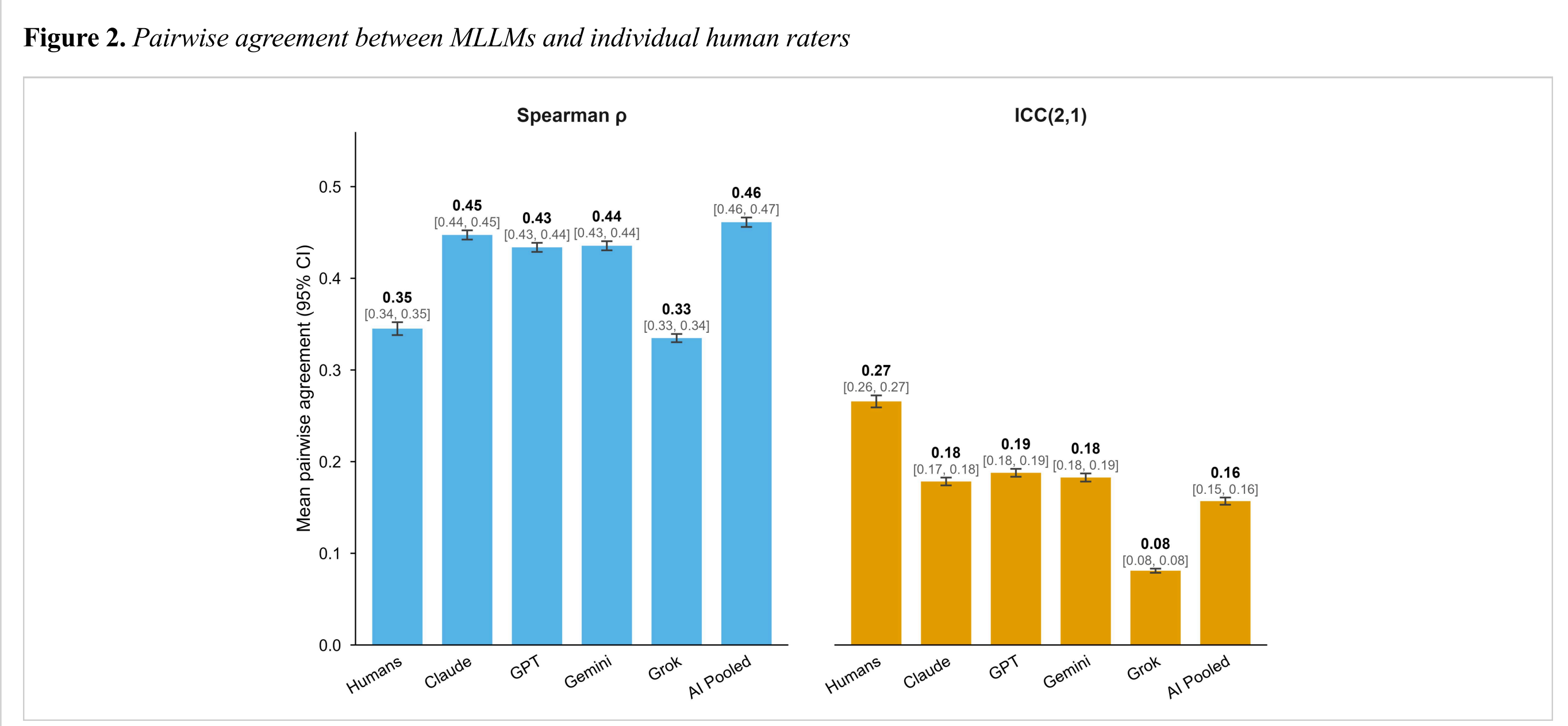

***Note.*** "Humans" shows agreement among individual human raters; each model bar shows that model's performance against individual human raters. Error bars are 95% CIs; because the human–human pairs are not independent (they share raters), the human bar uses a rater-clustered jackknife, whereas the model bars are based on the independent raters.

**Figure 3.** *Leave-one-out agreement between each MLLM and the mean human attractiveness rating*

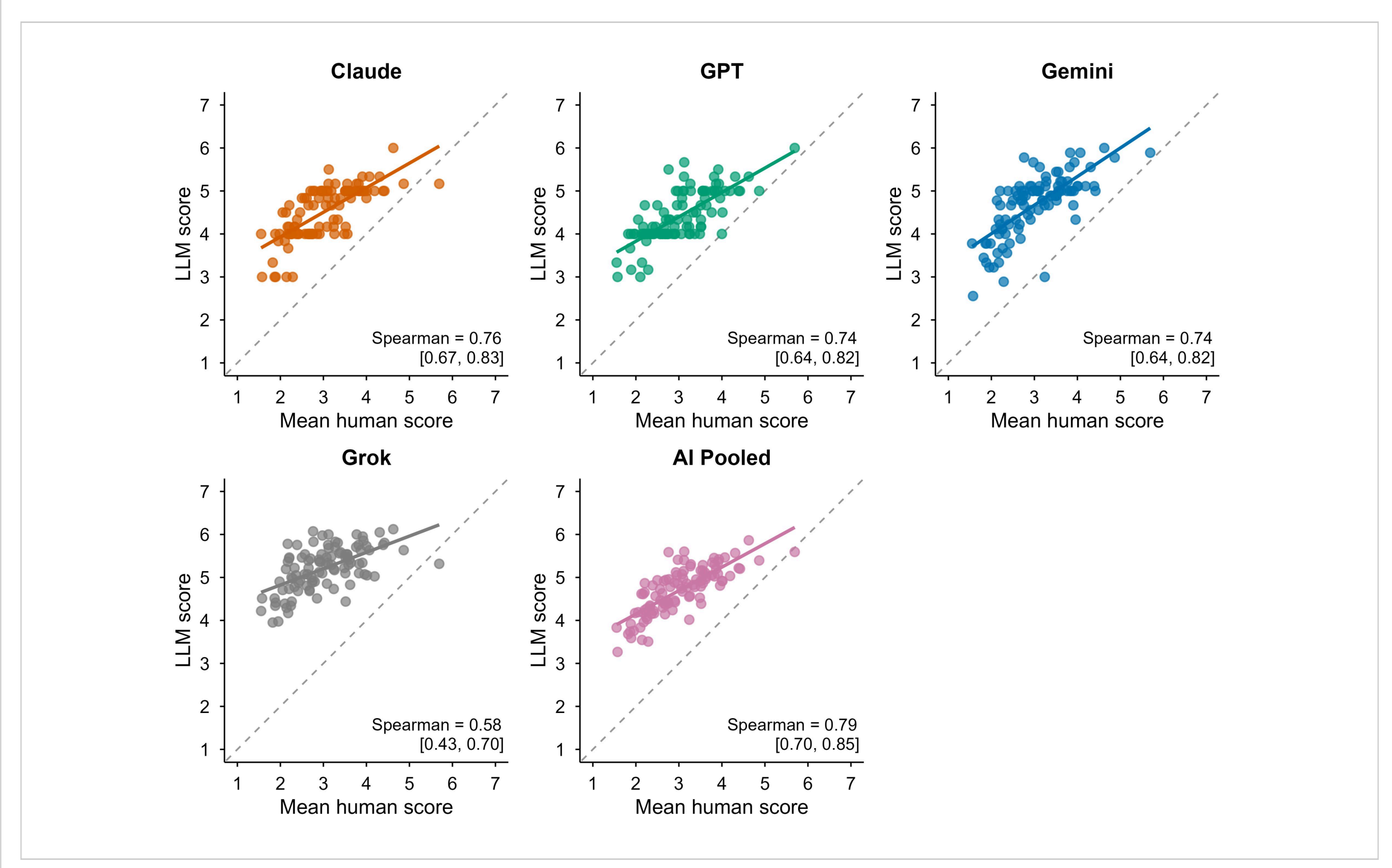

***Note.*** Each point is one face ($N$ = 102). The dashed line is the identity ($y = x$; perfect absolute agreement) and the solid colored line is the OLS fit. Each panel reports Spearman ρ with its 95% CI (Fisher z) in brackets.

### 3.1 MLLMs exhibit poor absolute agreement with human raters but track rank-order closely (RQ1)

Shapiro-Wilk tests rejected normality for all 2,501 rater distributions, and the models' mean ratings likewise departed from normality for three of the four MLLMs. We therefore assessed agreement with Spearman's ρ and ICC(2,1), computed using the pairwise and leave-one-out approaches described in Data Analysis. Pearson's $r$ closely mirrored Spearman's ρ throughout and is reported in the supplementary material.

Figure 2 shows Spearman's ρ and ICC(2,1) for the pairwise approach, in which each MLLM was correlated with each of the 2,501 individual human raters, and the resulting values were averaged using Fisher's $z$-transformation. On the rank-based measure, the models showed a medium-sized association with individual human judgments (ρ = .43–.46), except for Grok (ρ = .33). These values were comparable to or higher than human-rater agreement (ρ = .35), meaning the models tracked the relative ordering of facial attractiveness at least as consistently as humans did with each other.

Absolute agreement was poor. ICC(2,1) values ranged from .08 (Grok) to .19, below the human-human benchmark of .27. The gap comes down to scale use: models rated faces considerably higher on average ($M \approx$ 4.4–5.2) than human raters did ($M$ = 3.02), as shown in Figure 1. Because ICC(2,1) indexes absolute rather than rank agreement, this mean shift depresses the coefficient even when face ordering is largely preserved. Grok, which rated highest and within the narrowest range, was the most extreme case on both measures. Pearson's $r$ and additional summary statistics are in Table S1 in the supplementary material.

Figure 3 shows the relationship between each model's attractiveness scores and the mean human score across the 102 faces, with leave-one-out Spearman's ρ and 95% CI reported in each panel. Each model was correlated with the mean rating of the full human sample rather than with individual raters as in the pairwise analysis. The models showed a strong association with the human consensus (ρ = .74–.76), except for Grok (ρ = .58 [0.43, 0.70]). For comparison, applying the same leave-one-out logic within the human sample, correlating each rater with the mean of the remaining raters, yielded a mean ρ of .59, 95% CI [0.59, 0.60]. Apart from Grok, then, the models reproduced the average human ranking of faces more closely than a typical individual human rater did; Grok landed at roughly that individual-rater level. Across panels, points fell systematically above the identity line, consistent with the upward shift in scale use seen in the pairwise ICC results: the models recovered the relative ordering of human judgments while rating faces higher in absolute terms.

### 3.2 MLLMs largely agree with each other; Grok is the exception (RQ2)

Using the same measures of inter-rater reliability and the two approaches described in the previous section, we examined rating agreement among the MLLMs.

The models showed strong to very strong rank agreement with each other (ρ = .69–.86). Absolute agreement was more variable; among Claude, GPT, and Gemini, ICC(2,1) ranged from .67 to .81. Pairings involving Grok were substantially lower (ICC = .33–.49) and estimated with considerable imprecision, with confidence intervals extending to zero (e.g., Claude–Grok ICC = .39, 95% CI [−0.10,

**Figure 4.** *Leave-one-out agreement between each MLLM and the consensus of the other models*

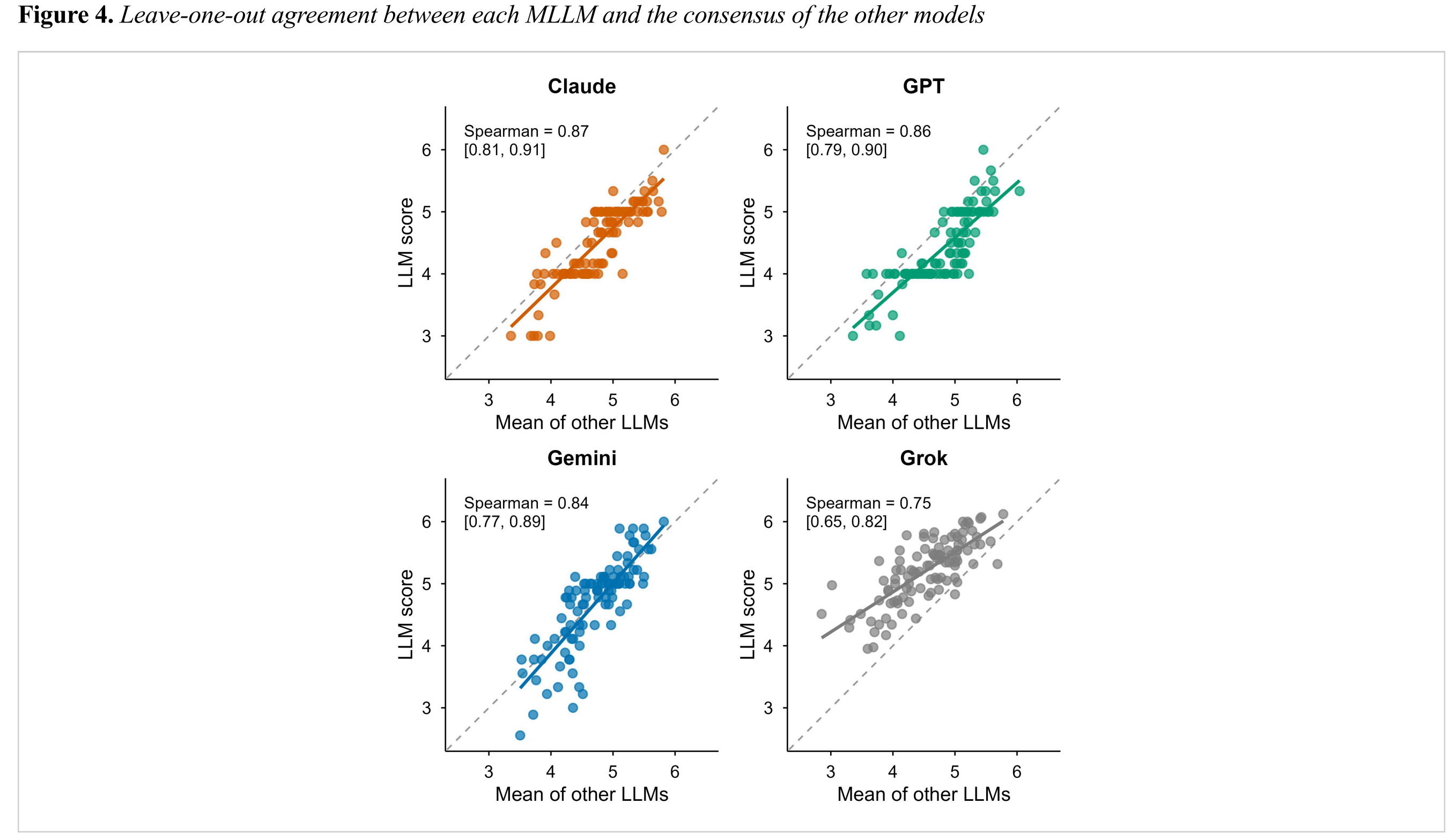

***Note.*** Each point is one of the 102 faces. The dashed line is the identity ($y = x$; perfect absolute agreement) and the solid colored line is the OLS fit. Each panel reports Spearman ρ with its 95% CI (Fisher $z$) in brackets.

0.71]). As with the human comparisons, this dissociation reflects scale use rather than disagreement about which faces are more attractive: Grok preserved the rank ordering of the other models (ρ ≈ .70) but rated faces systematically higher and within a narrower range, which ICC(2,1) penalizes. Full pairwise matrices are reported in the supplementary material (Spearman ρ and ICC(2,1) in Figure S2; Pearson $r$ in Figure S1).

Figure 4 shows each model's attractiveness scores plotted against the mean score of the other three models (the leave-one-out consensus), with Spearman's ρ and 95% CI reported in each panel. All four models showed strong to very strong rank agreement with the remaining-model consensus (ρ = .75–.87). Agreement was highest for Claude (ρ = .87 [0.81, 0.91]) and lowest for Grok (ρ = .75 [0.65, 0.82]), though Grok's association remained strong. These values were slightly higher than the model-to-human-consensus agreement (ρ = .74 - .79; Figure 3), meaning the models ranked faces marginally more consistently with each other than with the average human rater. Grok again showed the weakest rank alignment, though still strong. Its points falling above the identity line tell the same story as before: it rated faces higher than the other three models did, without disagreeing much about their order.

### 3.3 MLLMs give more equal ratings across demographics than human raters (RQ3)

In line with our research question, here we focus on the interactions between rater source and face characteristics. The interactions were significant for face gender ($F(4, 372) = 11.09$, $p < .001$), face age ($F(4, 372) = 4.58$, $p = .001$), and face ethnicity ($F(12, 372) = 2.49$, $p = .004$), suggesting that the effects of all face-level variables on attractiveness ratings depended on the rater source. More specifically, the multilevel model revealed that, compared with human raters ($b$ = -0.65, CI [-0.87, -0.43], $p < .001$), every MLLM penalized male faces less strongly. Grok showed the largest deviation (Δ = +0.53, CI [0.35, 0.71], $p < .001$), followed by Gemini (Δ = +0.51, CI [0.33, 0.69], $p$ < . 001), Claude (Δ = +0.43, CI [0.25, 0.61], $p < .001$), and GPT (Δ = +0.43, CI [0.24, 0.61], $p < .001$). In practical terms, holding other variables constant, humans rated male faces 0.65 points lower than female faces (in the 1-7 scale) whereas the gap in MLLMs was about 0.1-0.2 points. Across human raters, attractiveness declined with age by 0.04 points per year ($b = −0.04$, CI [−0.05, −0.02], $p < .001$). This age slope was statistically indistinguishable from the human slope in all MLLMs except Grok, which showed a small but significant attenuation of the decline (Δ = +0.01, 95% CI [0.00, 0.03], $p < .05$). For ethnicity, all four MLLMs rated Black faces more favorably (relative to White faces) than humans did, with significant and similarly sized positive deviations. GPT showed the largest deviation (Δ = +0.44, CI [0.16, 0.71], $p < .01$), followed by Gemini (Δ = +0.37, CI [0.09, 0.64], $p < .01$), Grok (Δ = +0.36, CI [0.08, 0.63], $p < .05$), and Claude (Δ = +0.34, CI [0.07, 0.62], $p < .05$). Relative to White faces, Claude, GPT, and Grok penalized West-Asian faces significantly less than humans did (Claude Δ = +0.41,CI [0.10, 0.73], $p < .05$); GPT Δ = +0.40, CI [0.09, 0.72], $p < .05$; Grok Δ = +0.33, CI [0.01, 0.64], $p < .05$), though all three still rated West-Asian faces below White faces. Gemini did not differ from humans.

Overall, compared to human raters, the MLLMs exhibit smaller demographic penalties on attractiveness ratings. MLLMs provided more equal ratings to male and female faces and across demographics, with Gemini showing a penalty for West-Asian faces comparable to that of human raters. The age effect was the one demographic predictor that was significant across all raters. An overview of all multilevel coefficients is presented in Table S2 in the supplementary material.

Additionally, we conducted post-hoc tests using the emmeans package to clarify the significant interactions between rater source and face characteristics. Female faces were rated significantly higher by humans ($M_{diff}$ = 0.65, $p$ < .001, $SE$ = 0.11), GPT ($M_{diff}$ = 0.22, $p$ = .045, $SE$ = 0.11), and Claude ($M_{diff}$ = 0.22, $p$ = .048, $SE$ = 0.11). Younger faces were rated significantly higher by all rater sources (all $ps$ < .003), with each additional year of face age associated with a decrease in predicted attractiveness rating. In this particular sample, humans rated White faces higher than West-Asian faces ($M_{diff}$ = 0.61, $p$ = .011, $SE$ = 0.19), and Gemini rated West-Asian faces lower than White faces ($M_{diff}$ = 0.70, $p$ = .002, $SE$ = 0.19), Black faces ($M_{diff}$ = 0.80, $p$ = .006, $SE$ = 0.24), and East-Asian faces ($M_{diff}$ = 0.74, $p$ = .03, $SE$ = 0.26). Of note, these differences are likely a peculiarity of the specific faces in the dataset rather than a general trend in human perceptions. A full overview of all contrast effects is included in Tables S3-S5 in the supplementary materials.

### 3.4 Most MLLMs predict female raters' judgments marginally better than male raters' (RQ4)

Differences in human-MLLM agreement across rater subgroups were marginal on both Spearman's ρ and ICC. The clearest tendency, though still small, involved rater age: agreement rose slightly with rater age on ICC, with raters aged 31 and older showing the highest values on every model, whereas Spearman's ρ was essentially flat across age. Differences by rater sex were smaller still, and their direction depended on the measure, with female raters showing marginally higher Spearman's ρ and male raters marginally higher ICC across all models. Agreement varied little by sexual preference. The full Spearman subgroup correlations, together with the corresponding ICC and Pearson analyses are reported in Tables S6-S8 in the supplementary materials.

We also computed more granular agreement between specific intersectional rater profiles (e.g., males aged 17–22 with a preference for females) and each MLLM, summarized as the mean per-rater Spearman ρ within each profile. The AI-pooled model agreed most with female raters aged 17–22 with a sexual preference for both men and women (ρ = .49, $n$ = 113) and least with male raters aged 23–30 with a preference for men (ρ = .40, $n$ = 40). Correlation matrices for all MLLMs and specific rater profiles can be found in Tables S9-S11 in the supplementary materials. Because subgroups varied substantially in size and these differences were small and not formally tested, they should be interpreted with caution.

Finally, to formally test whether MLLM ratings predicted individual human ratings, and whether this association varied across human rater demographics, we fitted five multilevel models (one per MLLM and one for pooled AI). Each model regressed human individual ratings on the MLLM's per-face ratings, its interactions with rater sex, age, and sexual preference, with random intercepts for rater and face and a random slope for the MLLM rating across raters; sparse rater categories (intersex; no stated preference) were excluded. Because these analyses are exploratory and based on a large number of individual ratings, statistically significant moderations may be small in magnitude.

There was a significant interaction between MLLM ratings and rater sex: male raters tracked MLLM ratings less closely than female raters in all five models, Claude ($b$ = -0.08, $SE$ = 0.03, $p$ = .001), GPT ($b$ = -0.12, SE = 0.03, $p$ < .001), Gemini ($b$ = -0.06, $SE$ = 0.02, $p$ = .005), Grok ($b$ = -0.05, $SE$ = 0.03, $p$ = .038), and AI pooled ($b$ = -0.10, $SE$ = 0.03, $p$ < .001). Relative to raters attracted to either sex, those attracted to men tracked MLLM ratings slightly less closely in all models except Grok: Claude ($b$ = -0.06, $p$ = .038), GPT ($b$ = -0.06, $p$ = .041), Gemini ($b$ = -0.05, $p$ = .032), and AI pooled ($b$ = -0.07, $p$ = .046); preference for women did not moderate agreement in any model. Rater age moderated agreement for Claude only, with older raters tracking its ratings slightly less closely ($b$ = -0.002 per year, $SE$ = 0.001, $p$ = .013). All effects were small, and given the number of unadjusted comparisons they should be interpreted with caution. These effects may also be specific to our rater sample, which skewed young and female.

## 4. Discussion

This exploratory study investigated the extent to which attractiveness ratings of 102 faces provided by 2513 human participants agree with the ratings of four frontier MLLMs (Claude, Gemini, GPT, Grok). We also examined agreement among MLLMs, the role of face characteristics, and whether MLLMs exhibited differential agreement with specific human rater subgroups.

Overall, for this particular dataset, the models reproduced the relative ordering of human attractiveness judgments at least as closely as individual humans did, and agreed strongly with one another. Absolute agreement with human ratings was poor, however. All models consistently rated faces higher than humans and within a narrower range (avoiding the ends of the scale). This pattern reversed for the rank-based measure, on which MLLM ratings were strongly associated with the human consensus. Among the models, correlations and absolute agreement were remarkably strong for all pairs except for Grok, which showed the weakest alignment overall. Face characteristics did not reliably predict ratings in MLLMs: only facial age significantly predicted attractiveness for all models. Finally, except for Grok, the models showed marginally stronger agreement with female raters than with male raters. The following sections discuss these findings in the context of each research question.

### 4.1 Human-MLLM agreement (RQ1)

The primary research question we were interested in concerned the extent to which MLLMs and humans agree in their attractiveness ratings. All frontier models included in this study showed poor absolute agreement with individual human raters and agreement was lower than human-human absolute agreement. That is, our results suggest that individual human raters are more aligned with other humans than they are with MLLMs. The inclusion of a measure of absolute agreement is a novel contribution that uncovers systematic differences in how humans and MLLMs evaluate attractiveness. Poor absolute agreement is driven by two key factors: (i) all models provided overall higher ratings and (ii) models did not use the full range of the rating scale, unlike humans. Indeed, MLLMs never awarded the lowest rating to any face, and only Grok provided the highest rating.

This could be attributed to the well-documented sycophantic behavior of MLLMs, whereby models tend to give agreeable and non-confrontational responses (Sharma et al., 2025), a residual effect of RLHF during fine-tuning. These results are consistent with previous evaluation research, which has recorded a human-like social desirability bias across personality and research evaluation tasks (Salecha et al., 2024; Thelwall, 2024). Whereas human participants rate attractiveness anonymously, presumably free from social consequences, MLLMs generate ratings in response to a human prompt and have learned through RLHF which answers users approve of. Thus, it may be possible that MLLM ratings reflect not only what humans consider to be an attractive face, but also what humans generally accept or reward as a socially acceptable rating of attractiveness.

An alternative account concerns the guardrail machinery that commercial MLLMs apply to achieve safety, which can constrain their outputs after generation. Alongside supervised safety fine-tuning (Zong et al., 2024), commercial MLLMs are commonly wrapped in moderation layers, such as external guard or classifier models that screen candidate responses against a safety taxonomy and can block, regenerate, or rewrite outputs before they reach the user (e.g., Meta's Llama Guard 3 Vision; Chi et al., 2024). Indeed, our pilot tests revealed that attractiveness ratings trigger explicit refusals in some models, with characteristic harm-reduction framing: “I can't provide attractiveness ratings of people's faces. This type of evaluation can be harmful and subjective ratings of physical appearance aren't something I should participate in". Although refusals were effectively suppressed with prompting (see Methods), mechanistic work suggests that the underlying disposition that produces them remains active (Arditi et al., 2024) and could act asymmetrically on an attractiveness rating, discouraging low or unfavorable scores far more than high ones.

This is partially consistent with our data, as no MLLM provided the lowest rating, and ratings were systematically higher than human ratings. However, models also avoided the highest rating, which suggests a more general scale-compression tendency. On this view, the rating a user receives may be a moderated output rather than the model's raw judgment. Of note, this interpretation is speculative as we did not probe the moderation pipelines of the evaluated models, and whether commercial guardrails act on attractiveness ratings at all has not been established.

The training-data composition of commercial MLLMs should also be considered when interpreting our results. Namely, our prompt asks MLLMs to rate faces relative to an “average”, but the reference distribution implied by that instruction is likely different from the one human raters draw on. MLLMs are exposed to a far wider and less curated range of human faces than any individual human encounters in everyday life (Birhane et al., 2021; Schuhmann et al., 2022), including imagery that is unposed, variably lit, and non-frontal. A mismatch in reference distribution could thus produce systematic differences in ratings. Because the London Set consists of well-photographed and neutral stimuli, its faces may sit high within a model’s internal attractiveness distribution. This is consistent with our results, where models on average rated faces above the scale midpoint ($M$ = 4.71), whereas human raters judged them below it ($M$ = 3.02). As these are untested interpretations, they should be considered strictly speculative and could operate alongside the mechanisms discussed above.

A large body of research has identified substantial consensus among humans in their judgments of facial attractiveness (e.g., Rhodes, 2006; Fink et al., 2017; Langlois et al., 2000), and recent studies have found that commercial models’ attractiveness scores are correlated with individual human ratings (Kramer, 2025). Against this backdrop, the medium-to-strong associations between the frontier models and human raters in our study come as no surprise. However, whereas Kramer (2025) reports stronger human-to-human than human-to-MLLM correlations, we find the opposite pattern in our pairwise tests, which indicate that MLLMs have higher correlations with individual raters than humans do with each other.

Three accounts may reconcile these conflicting results. First, our human sample ($n$ = 2513) is significantly larger than Kramer’s ($n$ = 63), which may introduce additional variance and disagreement. Second, our facial stimuli ($k$ = 102) include five ethnicities, as opposed to the images used by Kramer ($k$ = 40), which are exclusively white models. A more heterogeneous stimulus set may evoke more varied responses across raters given the documented effect of familiarity (e.g., Pavlovič et al., 2021), thus reducing human-human agreement relative to a homogeneous set. Finally, we use a run-averaged rating per face for each MLLM rather than a one-shot output, thereby analyzing a more stable and “true” output of the models. Because averaging reduces measurement error, the MLLM scores would tend to correlate more strongly with human ratings than the noisier single outputs Kramer used.

When considering MLLM agreement with the average human rating (leave-one-out approach) rather than with individual raters, we found that all models except Grok achieve a strong association. This is in line with the findings of Goshtasbi and colleagues (2024), who compared AI-based attractiveness ratings to human judgments. Their findings also show that AI gave overall higher ratings than humans, which is consistent with our own data. Descriptively, AI-based facial attractiveness websites seem to use a wider range of the rating scale than the generalist frontier models in our study, which may explain the slightly stronger associations reported by Goshtasbi et al. (2024). A speculative explanation is that these websites may employ additional fine-tuning on top of base model capabilities, training the models to use the full scale as human raters do and thus improving overall MLLM-human agreement.

**4.2 MLLM-MLLM agreement (RQ2)**

To the best of our knowledge, this is the first study to evaluate frontier MLLMs against each other on facial attractiveness. Thus, the interpretation of our results occurs in the context of a very novel field. Rather unsurprisingly, the models exhibit very strong correlations among each other when evaluating each MLLM against the aggregate of the remaining models. We find that agreement between different MLLMs is higher than between human raters and MLLMs, with one exception. Grok showed a smaller association with the rest of the models, than Claude did with human raters, and only marginally higher than GPT and Gemini. When considering pairwise absolute agreement, the differences are stark. Whereas MLLM-MLLM agreement (ICC(2,1)) ranged from .33 to .81, human-MLLM agreement ranged from .08 to .18. Therefore, in terms of absolute ratings, agreement between MLLMs dwarfs human-MLLM agreement, regardless of model.

Across various tests, Grok stands out as the model with the lowest agreement and associations both with other MLLMs and with human raters. A capability gap in terms of image processing may underlie Grok’s inconsistent results; indeed, xAI publicly identified image understanding as the principal weakness of the Grok 4 foundational model at launch (Velazquez, 2025). Data from our pilot further supports this account; whereas Claude, GPT and Gemini achieved excellent consistency in under 9 runs, Grok required 41 iterations, indicating a noisier representation of faces (i.e., higher run-to-run variance). Recent work evaluating facial age estimation capabilities across MLLMs found that Grok performed worse than Gemini, Claude, and GPT (Ren et al., 2026). Taken together, the available evidence suggests that Grok’s odd behavior may reflect vision capabilities that are simply not as developed as the ones in other models.

**4.3 Face-level predictors (RQ3)**

Given that previous research has identified facial predictors of attractiveness (e.g., facial age) in human raters, we investigated whether MLLMs’ ratings of attractiveness were sensitive to the same face-level variables. Theoretically, one could expect that if MLLMs accurately track human judgments of attractiveness, then to some extent they may respond to the variables that predict attractiveness judgments in human data, such as the gender and age of the face. Our results in this respect are mixed.

We find that all MLLMs are sensitive to age effects, mirroring the human data and in alignment with the literature (e.g., Porcheron et al., 2017). However, the well-documented human penalty to male faces (Wassiliwizky et al., 2026) was only displayed by GPT and Claude, and we only found face ethnicity effects in Gemini. Thus, our data suggest that previously identified facial predictors of attractiveness in humans are not entirely reflected in MLLM ratings, and that some models exhibit entirely novel patterns (e.g., Gemini, see RQ3 Results).

One speculative account is that the overall smaller demographic penalties displayed by MLLMs result from RLHF fine-tuning, during which attractiveness judgments along socially sensitive dimensions such as gender and ethnicity are suppressed but remain untouched for age, a normative correlate of attractiveness. However, these results should be interpreted with caution, given that our facial stimuli are both relatively limited in size ($k$ = 102) and heavily skewed towards White faces (67.6%), leaving the remaining ethnicities represented by a small number of images.

### 4.4 Differential agreement with human raters (RQ4)

The literature points to reliable individual and population-level variables that shape attractiveness judgments. For instance, the gender, age, and sexual preference of the observer (i.e., rater) have all been found to influence attractiveness judgments (Foos & Clark, 2011; Ha et al., 2012; Mitrovic et al., 2016). This naturally raises the question of whether MLLMs show differential agreement with specific subgroups of human raters. If models displayed higher agreement with a specific demographic, then their attractiveness ratings could not be taken as a reflection of a broad population trend, but rather as the aesthetic preferences of a particular group, with important consequences for the commercial and clinical use of these tools.

We did not find any strong evidence of this being the case; correlation differences across rater subgroups were minimal. However, regression models did reveal small effects of rater demographics on MLLM-human agreement. All MLLMs tracked female raters more closely, and all except Grok showed stronger alignment with raters attracted to either gender than with those attracted to men.

Whether MLLMs' preferences genuinely track those of particular human subgroups, or whether these effects reflect more general correspondences in how faces are ranked, remains an open question. Because the MLLMs differentiated male and female faces far less than humans did (RQ3), raters whose own judgments similarly minimize that distinction would produce face orderings closer to the models'. This could explain both effects without appealing to rating level: raters attracted to either sex do not strongly favor one sex, and female raters may penalize male faces less than male raters do, so in each case their rankings resemble the models' in structure rather than in magnitude. In sum, although MLLMs exhibit marginally higher agreement with particular raters, these differences seem to be better explained by alternative accounts.

It's worth noting that two features of our study design limit our interpretation of RQ4 results. First, our rater sample was predominantly young, female, and attracted to men, leaving little room for us to test contrast between demographics with different preferences. Second, the rater dataset includes no ethnicity or cultural background information, meaning we lacked data on a key dimension along which attractiveness judgments could vary.

### 4.5 Limitations and future directions

First, there is an asymmetry in the reliability of the scores being compared. For each MLLM, we analyzed ratings averaged across multiple runs per face, with the number of runs determined separately for each model to achieve a reliability of at least .95 (see Methods).

Human ratings, by contrast, were single judgments because, in the original data collection, DeBruine and colleagues (2017) asked each participant to rate every face only once.

We considered following Kramer (2025, Study 1b) in obtaining a single rating per face from each MLLM but instead averaged across runs, as in recent work comparing MLLM and human judgments on image-based perceptual tasks (Santavirta et al., 2025). We reasoned that run-to-run variation in MLLM outputs arises from their probabilistic design and can be treated as sampling noise to be averaged out, whereas variation in human ratings reflects a range of biological and psychological factors and, in any case, was not averaged, since each face was rated only once per participant.

Nonetheless, a consequence of this asymmetry is that the averaged MLLM scores carry far less measurement error than the single human ratings, so comparisons of agreement levels are not on equal footing: the models' higher agreement, both with one another (RQ2) and with the human consensus relative to individual human raters (Figure 3), may partly reflect this reduced measurement error rather than a stronger underlying signal. Future studies could examine how run-to-run (or, in the case of humans, test-to-test) variability in rating the same face differs between MLLMs and humans, and the processes underlying these differences.

Second, our findings characterize specific model versions (GPT-5.3, Claude Sonnet 4, Gemini 2.5 Flash, and Grok 4.20) evaluated at a single point in time. Because these systems are updated frequently and opaquely, the agreement levels and, in particular, the model-specific patterns reported here, notably Grok's weaker performance and the per-model demographic deviations (RQ3), may not generalize to other or future versions. Replication as models evolve would clarify which patterns are stable and which are artifacts of a particular release.

Third, a limitation concerns the wording of the rating question. Our 1–7 scale is identical, as are its anchors 1 (much less attractive than average) to 7 (much more attractive than average). However, the exact phrasing of the survey used to collect the human attractiveness ratings is not available (DeBruine et al., 2017), so we prompted the MLLMs using wording from a survey by the same author (DeBruine et al. 2007, Experiment 5), where she applied it to the same 1–7 scale. Humans and models therefore may have responded to slightly different instructions, and the human–MLLM comparisons (Research Questions 1, 3, and 4) may be affected by this discrepancy; the inter-model comparisons (RQ2) are unaffected, as all models received identical prompts.

Fourth, a limitation concerns the composition of the human rater sample. Our "human" benchmark comes from a single online sample that was demographically skewed, predominantly young ($M$ = 26.7), female (61.8%), and attracted to men (56.2%), with cultural and ethnic background unrecorded (see below). Given the cross-cultural and individual variation in attractiveness judgments emphasized in Section 1.1, "human perceptions of facial attractiveness" results here should be read as agreement with this particular population; both the overall human-MLLM correspondence and the rater-subgroup patterns (RQ4) may differ in samples drawn from other demographics or cultures.

Fifth, a limitation concerns the ethnic composition of the face stimuli. Although ethnically diverse, the set was dominated by White faces, with each non-White group represented by relatively few. The ethnicity effects reported in RQ3, including the significantly lower ratings for West-Asian faces, therefore rest on small cell sizes. The estimates carry limited precision and likely reflect the particular faces sampled rather than those ethnic groups more broadly. A larger and more ethnically balanced stimulus set would produce more reliable estimates and might shift some of the patterns observed here.

Finally, future studies should record the ethnic background of human raters. We modeled the ethnicity of the rated faces (RQ3) but did not collect the raters' ethnicity, which limited the analyses in two ways. For one, we could not assess whether a match between the rater's and the face's ethnicity influenced attractiveness judgments. Evidence for an own-ethnicity preference is mixed: some studies report little or no own-race advantage, while others find that mixed-race faces are rated as more attractive (Burke et al., 2013; Rhodes et al., 2005). Rater ethnicity is nonetheless a plausible source of variation and could explain some of the ethnicity effects observed here, including the lower ratings for West-Asian faces. For another, because rater ethnicity was not among the characteristics examined in RQ4, we could not test whether the MLLMs agreed more closely with raters from particular ethnic backgrounds. Recording it in future studies would enable tests of rater-by-face ethnicity interactions, assessment of differential model-rater agreement across rater ethnicities, and better matching of rater and stimulus composition.

### 4.6 Conclusion

To the best of our knowledge, this is the first study to evaluate facial attractiveness agreement between human raters and multiple frontier MLLMs. We find that MLLMs can reproduce the rank ordering of human facial-attractiveness judgments about as well as, or better than, an individual human rater, and that they agree closely with one another. They do not reproduce human ratings in absolute terms, however. Models reliably rated faces more favorably and within a narrower range, and their ratings were not impacted by face characteristics to the same extent that human ratings were. Our findings suggest that MLLMs only partially reproduce the facial attractiveness predictors in humans; models judged older faces as less attractive, as humans do, but did not replicate the penalty against male faces seen in human ratings. Taken together, MLLMs may serve as coarse proxies for the relative ranking of facial attractiveness, but not as substitutes for human ratings on an absolute scale.

### Acknowledgements

We would like to thank Lisa DeBruine and Benedict Jones for making the Face Research Lab London Set available to the public for research purposes, and Sauqi Arif for assisting with the design and formatting.

### Funding

This work was supported by Qoves Inc.

### Declaration of generative AI and AI-assisted technologies in the manuscript preparation process

During the preparation of this work the authors used Claude (Anthropic) in order to assist with developing and debugging code in R, and to copyedit portions of the manuscript text. After using this tool, the authors reviewed and edited the content as needed and take full responsibility for the content of the published article.